\documentclass[letterpaper, 10 pt, conference]{IEEEtran}
\IEEEoverridecommandlockouts
\usepackage{etextools}
\usepackage{amssymb}
\usepackage{float}
\usepackage{makeidx}
\usepackage{amsmath}
\usepackage{amsfonts}
\usepackage{bbm}
\usepackage{epsfig}
\usepackage{epsf}
\usepackage{psfrag}
\usepackage{verbatim}
\usepackage{color}
\usepackage{multirow}
\usepackage{tabularx}
\usepackage{booktabs}
\usepackage[tight,footnotesize]{subfigure}
\usepackage{array}
\usepackage{soul}
\usepackage{footnote}
\usepackage{cite}
\usepackage{dblfloatfix}

\usepackage{color, colortbl}
\usepackage{graphicx}
\graphicspath{{Figures}}
\DeclareGraphicsExtensions{.pdf,.png}
\usepackage{bigstrut}
\usepackage[english]{babel}

\usepackage{csquotes}

\usepackage[T1]{fontenc}
\usepackage{algorithm, setspace}
\usepackage{algpseudocode}
\usepackage{url}
\usepackage{multirow}
\usepackage{float}
\usepackage{xcolor}

\newcommand{\p}[1]{\smallskip \noindent \textbf{{#1}.}}
\newcommand{\eq}[1]{Equation~(\ref{eq:#1})}
\newcommand{\fig}[1]{Figure~\ref{fig:#1}}
\usepackage{balance}
\let\labelindent\relax
\usepackage{enumitem}

\usepackage{pgfplots}
\pgfplotsset{compat=1.18}
\usepgfplotslibrary{groupplots}   
\usepackage{pifont}
\providecommand{\cmark}{\ding{51}}
\definecolor{oursblue}{RGB}{42,143,189}
\definecolor{gmorange}{RGB}{255,153,0}

\usepackage[colorlinks=true,
            linkcolor=orange,
            citecolor=orange,
            urlcolor=orange,
            filecolor=orange,
            hypertexnames=false]{hyperref}

\usepackage{makecell}            
\begin{document}
\title{\LARGE
GraspTwin: Zero-Shot Task-Oriented \\ Grasp Optimization via a Digital Twin
\vspace{-0.2em}
}

\author{Daniel J. Evans$^1$, Yinlong Dai$^1$, Simon Stepputtis$^2$, and Dylan P. Losey$^1$
\thanks{This work is supported in part by NSF Grant $\#2337884$.
}
\thanks{$^1$\href{https://collab.me.vt.edu}{Collab}, Dept. of Mechanical Engineering, Virginia Tech, Blacksburg, VA 24061. \texttt{\{danielevans, daiyinlong, losey\}@vt.edu}}
\thanks{$^2$\href{https://tealab.ai}{TEA Lab}, Dept. of Mechanical Engineering, Virginia Tech, Blacksburg, VA 24061. \texttt{stepputtis@vt.edu}}
}

\maketitle


\begin{abstract}

As robots transition from structured factory settings into homes, they are required to interact with an ever-increasing variety of objects. 
Many tasks require \textit{grasping}, and often it is not sufficient to just pick up the target object.
Consider a task like ``pouring coffee'' --- to facilitate the subsequent pouring, the robot should grasp the mug by its handle.
Existing learning-based approaches for grasping either find robust and collision-free grasps that are largely agnostic to the task (e.g., picking up the mug by its rim), or leverage foundation models to propose task-appropriate grasp locations that lack fine-grained physical grounding (e.g., reaching for and missing the handle). 
In this work, we bridge these approaches with a \textit{real-to-sim-to-real} framework.
Based on a single RGB-D observation, we construct a digital twin of the environment, query a large foundation model to propose grasps that align with the object's affordances and task description, and then optimize the proposals to ensure robustness and plausibility before executing the result on the real robot. 
Our key insight is that the grasp proposals of the foundation model should be regarded as semantic priors that serve as seeds for local, gradient-free optimization.
We leverage Bayesian optimization with Thompson sampling to draw batches of nearby poses, which are subsequently evaluated in parallel under domain-randomized physics rollouts. 
The resulting grasp is \textit{both task-oriented and physically feasible} for execution by the robot arm.
Our full zero-shot real-world transfer only takes a few minutes and improves task-oriented grasping success by up to 33\% as compared to other state-of-the-art pipelines. Our code is available here: \\ {\footnotesize\url{https://github.com/VT-Collab/GraspTwin/}}

\end{abstract}


\section{Introduction} \label{sec:intro}

Robot arms are moving towards generalized manipulation policies.
One integral part of these policies is the \textit{grasp} --- the way that the robot picks up and holds objects.
By itself, grasping challenges the robot's ability to find a valid, fine-grained configuration in which it can hold the target object without accidentally colliding with other scene objects.

\begin{figure}[t]
    \centering
    \includegraphics[width=1.0\linewidth]{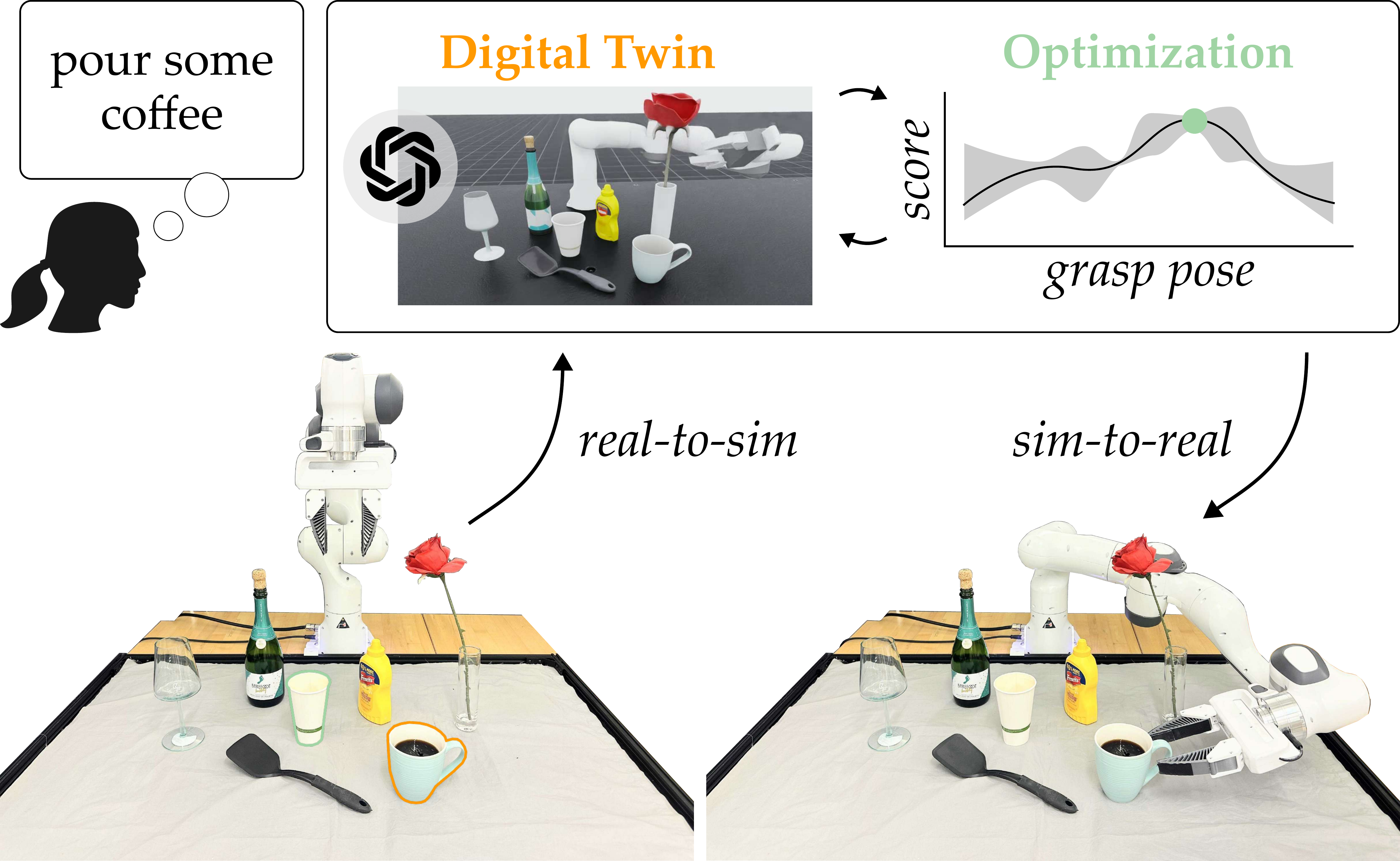}
    \caption{We present \textbf{GraspTwin}, a zero-shot method for identifying grasps that are both task-oriented and physically feasible. Given a single RGB-D image of the scene, GraspTwin constructs a digital twin. Within that digital twin we query foundation models to propose grasps that facilitate the downstream task. These semantically-aware grasps seed local optimization while enforcing physical constraints (e.g., unintended collisions) and domain randomization. The resulting grasp pose is then executed by the real arm.}
    \label{fig:front}
    \vspace{-1.8em}
\end{figure}

What makes this even more challenging is that \textit{not all grasps are created equal}.
Imagine you tell a robot arm to ``pour some coffee'' (see \fig{front}).
In general, a robust place to grasp a coffee mug is the body, and so a naive robot might select this grasp pose.
But when it moves to ``pour the coffee,'' the gripper fingers are inside the mug --- causing coffee to splash across the gripper.
The correct grasp here is \textit{task-oriented}.
It reasons about the object's affordances and the downstream task, and selects a grasp pose that anticipates these manipulation requirements.
Of course, the resulting grasp must be \textit{physically feasible}: preventing unintended collisions (e.g., knocking over the coffee) and ensuring robustness (e.g., holding the mug so it does not slip).

Existing works have developed multiple tools for general-purpose manipulation grasps.
At one extreme are learning-based approaches that use thousands of grasping attempts to train \textit{grasp-specific neural networks} \cite{tang2023graspgpt, tang2025foundationgrasp, deshpande2025graspmolmo}.
These approaches are excellent at finding and executing robust grasps (e.g., picking up the object) --- but they do not always consider whether those grasps are aligned with the task (e.g., picking up the object in the right way).
At the other end of the spectrum are \textit{language-based approaches} that apply foundation models to reason over the objects and desired task.
This includes explicitly reasoning about attributes and affordances~\cite{li2024shapegrasp, cui2026contact}, generating interaction hypotheses~\cite{tang2026grasp,chen2024vlmimic, ye2023affordance}, or selecting interaction points in the input image~\cite{deshpande2025graspmolmo, yuan2024robopointvisionlanguagemodelspatial}.
While these language-based methods are effective at understanding the right type of grasp, they struggle to ground high-level decisions in embodied motion, often resulting in lower success rates despite their accurate proposals. 

To this end, we introduce a bridge between the \textit{reasoning} capabilities of foundation models and the \textit{grounding} capabilities of data-driven methods. 
Given a natural language task description, our goal is to tractably (e.g., in roughly 5 min) perform a semantic grasp on a new object without any in-context demonstrations or fine-tuning.
Instead of relying on trial-and-error in the real world, our insight is that:
\begin{center}
    \vspace{-0.3em}
    \textit{Robots can generate digital twins of common objects from a single depth image, and then optimize semantically-aware grasps within those simulated environments before performing the resulting grasp on the real robot.}
    \vspace{-0.3em}
\end{center}
We apply this insight to develop GraspTwin, a \textit{real-to-sim-to-real} pipeline for grounding semantic grasping (see \fig{front}).
Given the task description and the observation from a fixed RGB-D camera, we use vision-language models (VLMs) to propose grasps.
We then refine these semantically-aligned grasps within an autonomously constructed digital twin.
The optimized grasp captures the semantics of the object and task (e.g., holding the coffee by the handle), while maximizing grasp robustness when executed by the robot arm (e.g., taking into account kinematics and other potentially obstructing objects in the scene).
This ultimately allows the system to find grasps that are task-aligned and physically plausible.

Overall, we make the following contributions:

\p{Generating Semantic Proposals}
We first go from real-to-sim by building a digital twin from real-world observations.
We leverage the simulated object and robot to render views of different grasp proposals, and then query a VLM to assess these options.
The VLM ultimately samples grasps that align with the downstream task and object affordances.

\p{Optimizing for Robustness \& Feasibility}
We treat the VLM proposals as the seeds for local optimization.
More specifically, we apply Bayesian optimization with Thompson sampling to identify nearby grasps, and then we test these grasps in the digital twin under domain randomization.
Our end-to-end process from initializing the task to actually performing the grasp takes roughly 5 minutes on average.

\p{Demonstrating Zero-Shot Generalization}
We compare our approach to ablations and state-of-the-art baselines across 30 standardized tasks.
These tasks include clutter, objects placed on top of one another, and variations in lighting or camera position.
Across our simulations and real-world experiments, we find that our real-to-sim-to-real approach increases both absolute and semantically-aware grasping success.
\section{Related Works} \label{sec:related}

Our approach combines the task-oriented grasp proposals from foundation models with the task-agnostic optimization of stable grasps through a real-to-sim-to-real pipeline. 
Accordingly, we briefly survey each of these fields below.

\subsection{Task-Agnostic Grasp Generation} \label{sec:r1}
Common approaches to grasping known objects center around force-closure methods using analytic quality metrics~\cite{miller2004graspit}.
When the objects are unknown \textit{a priori}, methods such as~\cite{ten2017grasp} derive grasp candidates directly from point clouds using hand-crafted geometric features. 
Learning-based methods like GraspNet~\cite{fang2020graspnet} and Contact-GraspNet~\cite{sundermeyer2021contact} replace these heuristics with neural networks trained on large grasp datasets.
In practice, these methods derive 6-Degree-of-Freedom (DoF) grasps from partial point clouds of the target objects.
AnyGrasp~\cite{fang2023anygrasp} extends this framework to dense and temporally consistent detection in clutter, while GraspGen~\cite{murali2025graspgen} models the grasp distribution with a diffusion prior and learned discriminator.
The data used to train these grasp models was historically collected on robots, but recent works also applied foundation models to learn grasping at a large scale from human videos~\cite{wu2026human}.
Overall, task-agnostic approaches provide a robot-aware method for identifying feasible and stable grasps.
But these methods alone do not consider the \textit{semantics} of those grasps, i.e., which grasps are aligned with the current task.

\subsection{Task-Oriented and Semantic Grasping} \label{sec:r2}
In task-oriented grasping the goal is to not only select a stable grasp, but also a grasp that facilitates task completion~\cite{murali2021same}.
This generally requires the common sense reasoning of foundation models.
Some works have therefore used grasping datasets to fine-tune large foundation models \cite{tang2023graspgpt, tang2025foundationgrasp, deshpande2025graspmolmo, park2026learning}.
The resulting models are then applied to recognize object relationships \cite{tang2025foundationgrasp} or directly predict task-aligned grasps \cite{deshpande2025graspmolmo}.
Of course, this grasp-specific fine-tuning process requires a large labeled corpus, and the resulting model may lose some of its generalization capabilities.

Instead of training specialized language models, another approach is to simply apply off-the-shelf large language or vision models.
LERF-TOGO~\cite{rashid2023language} grounds language in a radiance field to localize the object's most suitable grasp pose.
Lan-Grasp~\cite{mirjalili2023lan} queries an LLM to determine which part to hold, and then asks a VLM to localize that part.
Similarly, ThinkGrasp~\cite{qian2024thinkgrasp}, AffordGrasp~\cite{tang2025affordgrasp}, ShapeGrasp~\cite{li2024shapegrasp}, and SegGrasp~\cite{li2024seggrasp} reason about affordances and object parts.
GraspDreamer~\cite{tang2026grasp} synthesizes human demonstrations with video generation models and then re-targets them to robots. 
Across all of these approaches, language models are used to answer the high-level question of \textit{where} to grasp.
In order to actually determine \textit{how} to perform that grasp, these methods typically pass the target region to one of the task-agnostic samplers outlined in Section~\ref{sec:r1}, which may change the grasping point.
This leads to a disconnected framework that \textit{separates the grasp proposal from its execution} --- leading to downstream errors where the proposed grasp is kinematically infeasible, collides with nearby objects, or misses functional constraints~\cite{taunyazov2023grace}.
By contrast, our method closes the loop between proposal and execution by iteratively testing and refining the task-oriented grasp in a physics simulator.

\subsection{Real-to-Sim-to-Real for Manipulation} \label{sec:r3}
The technical advance that makes our approach possible is the advent of real-to-sim generators that autonomously build simulated environments on the fly. Early work in this direction used generators to automatically synthesize novel simulation assets~\cite{zhou2023learning}, and subsequent systems extended this to constructing sim-ready scenes directly from real-world observations~\cite{dai2024automated}.
BundleSDF~\cite{wen2023bundlesdf} tracks and reconstructs previously unseen objects from RGB-D videos, while methods such as SAM-3D~\cite{sam3dteam2025sam3d3dfyimages} and Hunyuan3D~\cite{zhao2026hunyuan3d20scalingdiffusion} create textured meshes from a single image.
End-to-end approaches like RialTo~\cite{torne2024reconciling} and SimFoundry~\cite{ranawaka2026simfoundry} turn a scan or video of a real environment into a sim-ready digital twin with full physics and collisions enabled.
Our research applies these tools to create a \textit{real-to-sim-to-real} pipeline for task-oriented grasping: the robot takes a real image, generates a physics-based simulation, optimizes for a grasp with domain randomization, and then executes the result in the real world.

\section{Problem Statement} \label{sec:problem}

We consider the problem of task-oriented grasping in which a single robot arm must grasp an object while accounting for its current task.
The robot is given a natural language task description $\tau$ (e.g., ``scoop the food'') and an image $\mathbf{I}$ of its table-top environment.
Based on the task and image, the robot must identify the target object and find a suitable grasp pose $\mathbf{T} \in SE(3)$.
A successful grasp pose should not only reason about the task (e.g., gripping the spoon by its handle), but also the robot's kinematic constraints (e.g., orienting the grasp so that the robot can rotate the spoon) and potential collisions with other objects (e.g., accidentally knocking over a bowl).
Our method addresses these challenges by introducing a \textit{digital twin} that models the environment,  then proposing and optimizing a task-oriented grasp within it.

\p{Observations}
Our setting consists of a table with a mounted 7-DoF robot arm and a 1-DoF parallel jaw gripper.
The robot observes its environment using an external, fixed camera that provides RGB-D images $\mathbf{I} \in \mathbb{R}^{1920\times 1080 \times 4}$.
Let the robot's \textit{observation} $o \in \mathcal{O}$ include both its own joint $\mathbf{j} \in \mathbb{R}^7$ and gripper state $g \in \mathbb{R}$ along with the camera image $\mathbf{I}$.
For the purposes of our method, we will assume access to a model of the robot's kinematics and dynamics.
The robot arm can use these kinematics and dynamics to execute waypoint-based control and move along a joint-space trajectory.

\p{Objects}
On the table are a variety of household items.
We restrict our problem setting to only rigid objects that do not require articulation to grasp.
For example, if a pair of scissors is open on the table, we do not consider scenarios where we need to first close the scissors before picking them up.
The objects themselves can be crowded together or in contact with one another, provided they are not fully occluded from the camera.
Let the set of objects be denoted as:
\begin{equation} \label{eq:P1}
    \theta = \{(c_1, g_1, p_1), \ldots, (c_N, g_N, p_N)\}
\end{equation}
where $N$ is the total number of objects.
The $i$-th object is defined by its textual label $c_i$, its geometry $g_i$, and its pose $p_i$.
The robot does not have access to $\theta$: instead, the robot must try to infer whatever components of this object information are necessary to perform the grasp.

\p{Grasp Planner}
The grasp planner is a mapping from the user's task description $\tau$ (e.g., ``scoop the food'') and the robot's observation $o$ to a grasp pose $\mathbf{T} \in SE(3)$:
\begin{equation} \label{eq:P2}
    \mathbf{T} \sim \pi_{grasp}(\cdot \mid o, \tau)
\end{equation}
In our experiments, pose $\mathbf{T}$ consists of the desired grasp position and orientation --- we apply waypoint-based control to make the arm reach this pose.
We emphasize that this planner $\pi_{grasp}$ is not meant to complete the entire task $\tau$.
Instead, the robot uses its nominal policy $\pi$ to approach the object, switches to $\pi_{grasp}$ to perform the grasp, and then returns to $\pi$ after the grasp is complete.

\p{Objective}
Our goal is to obtain a grasp planner that is physically stable, does not collide with nearby objects, and is semantically meaningful for the downstream task.
This planner should take an RGB-D image of the scene and output a grasp pose that robustly holds the object while facilitating the current task.
Importantly, the robot does not have access to in-context demonstrations or real-world training attempts: the robot should grasp the object on its first trial.

\section{Optimizing Task-Oriented Grasps} \label{sec:method}

\begin{figure*}
    \centering
    \includegraphics[width=1\linewidth]{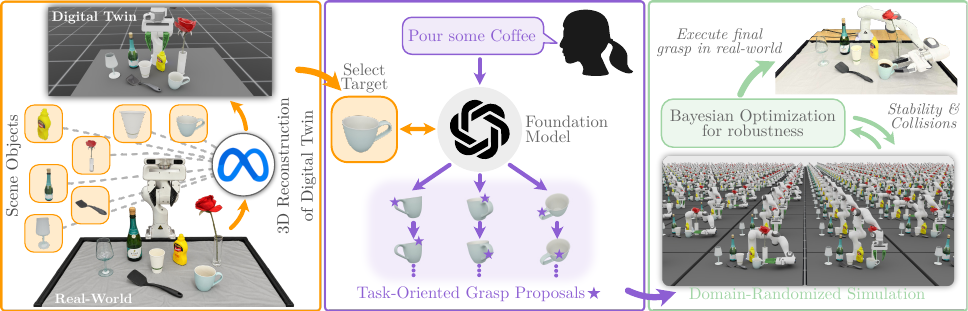}
    \caption{Overview of \textbf{GraspTwin}. (Left) We convert from real to sim by constructing a digital twin from an RGB-D image of the robot's environment. We estimate the scene objects $\hat{\theta}$ and identify $\hat{\theta}_1$, the target object for grasping. (Middle) We then render the target object $\hat{\theta}_1$ from multiple perspectives, and use a VLM conditioned on the task $\tau$ to select task-oriented grasp positions $\mathbf{P}_{\mathrm{VLM}}^{(i)}$ and an approach direction $\mathbf{R}_{\mathrm{VLM}}$. (Right) These proposals become the initial guesses for localized grasp optimization. We simulate grasp poses $\mathbf{T}$ in the digital twin with object perturbations $\Delta \theta$. Batch Bayesian optimization with Thompson sampling enables parallel evaluations for gradient-free optimization. The minimum-cost grasp pose $\mathbf{T}^*$ is executed by the real robot.}
    \label{fig:method}
    \vspace{-1.8em}
\end{figure*}

Our proposed method for zero-shot task-oriented grasping is outlined in \fig{method}.
We divide this method into three sequential stages.
First, we construct a digital twin of the grasping environment that focuses on rendering the target object and its surroundings (\textit{Real-to-Sim}, Section~\ref{sec:M1}).
Second, we leverage the common sense capabilities of VLMs to propose a set of grasp positions and orientations that are aligned with the given task (\textit{Task-Oriented Initialization}, Section~\ref{sec:M2}).
Finally, we perform gradient-free grasp optimization within the digital twin.
This optimization is seeded by the VLM-proposed grasps, and iteratively increases how robustly these grasps hold the object while remaining physically feasible and collision free (\textit{Grasp Optimization}, Section~\ref{sec:M3}).
The pose $\mathbf{T}^*$ output by the optimization is then performed directly on the robot (\textit{Sim-to-Real}): the entire end-to-end process is completed in roughly 5 minutes on average.

\subsection{Constructing the Digital Twin} \label{sec:M1}

Our first step is generating a physics-based digital twin from a single RGB-D image of the scene.
This digital twin will have two main applications.
First, it should reconstruct the unseen parts of the objects to create 3D meshes that can be viewed from multiple angles.
This will be essential for the VLM task-oriented grasp proposal --- to select what aspect of the object to grab, the VLM must view the entire object.
Second, the digital twin should serve as a testbed for different grasps.
If the digital twin is sufficiently accurate, then we can predict which grasps will likely succeed or fail, and leverage these results to optimize the grasp pose.
Hence, we have two key requirements for the digital twin.
The digital twin must (a) recreate complete object meshes from a single image; and (b) render robot-object interactions with sufficient fidelity so that grasps which work or fail in the twin similarly work or fail in the real world.

Recent advances in real-to-sim methods satisfy both of these requirements.
Given a single RGB-D image $\mathbf{I} \in \mathbb{R}^{1920\times 1080 \times 4}$ of the table-top setting, we first identify which object the robot arm should grasp.
This is a two-step process: we (a) use a VLM with an open-vocabulary classifier \cite{liu2024grounding} as a fallback, to detect and label objects in $\mathbf{I}$, and then (b) query an LLM conditioned on the task description $\tau$ and the labeled objects.
Once the target object is found, we next apply SAM-3D~\cite{sam3dteam2025sam3d3dfyimages} to convert the RGB-D image $\mathbf{I}$ into an estimate of the object set:
\begin{equation} \label{eq:M1}
    \hat{\theta} = \{\hat{\theta}_1, \ldots \hat{\theta}_N\} = \{(\hat{c}_1, \hat{g}_1, \hat{p}_1), \ldots, (\hat{c}_N, \hat{g}_N, \hat{p}_N) \}
\end{equation}
We assign the first index $\hat{\theta}_1 \in \hat{\theta}$ to the target object, and the following indices are sorted by their Euclidean distance from $\hat{\theta}_1$ (i.e., the pose of $\hat{\theta}_2$ is closest to the pose of $\hat{\theta}_1$).
For each object, $\hat{c}_i$ is the estimated label, $\hat{g}_i$ is that object's reconstructed 3D mesh, and $\hat{p}_i$ is the estimated pose.
We recognize that the robot only needs highly accurate models of the target object and its immediate surroundings.
Accordingly --- to accelerate the digital twin --- we monotonically decrease the mesh resolution of $\hat{g}_i$ as the index $i$ increases.

Now that we have an estimate of the objects $\hat{\theta}$, we can use the object meshes and poses to update a digital twin.
In our experiments we leverage Isaac Lab~\cite{mittal2025isaaclab}.
This physics-based digital twin is initialized with the robot arm mounted on a table, and the table is then populated with the extracted objects.
These objects are assigned dynamic parameters (e.g., mass, inertia, friction) which may not be accurate, but allow for physical interaction between objects and the robot.
As discussed in Section~\ref{sec:M3}, we will leverage domain randomization to increase our robustness to these incorrect dynamics.

\subsection{Identifying Task-Oriented Grasps} \label{sec:M2}

Equipped with a digital twin, the next step considers the robot's grasp itself.
The appropriate way to grasp an object depends on the task --- and so we start by identifying task-oriented grasp candidates before optimizing their fine-grained values.
Our methodology builds on related works~\cite{li2024shapegrasp,rashid2023language} that leverage the reasoning capabilities of off-the-shelf VLMs.
What makes this aspect of our approach unique is that we have access to a digital twin and reconstructed object meshes.
We leverage these tools to render 2D images of the target object and scene from multiple perspectives, and then query a VLM to locate where and how the robot should grasp the target object within these images.

\p{Chain-of-Thought Grasp Description}
To \textit{steer} the outputs of our VLM, we first prompt the model to describe a grasp that facilitates task $\tau$.
For example, consider ``pour coffee.''
This instruction does not define which part of the cup should be grasped, or how the robot needs to hold that cup to complete its pour.
We therefore prompt the VLM to reason across $\tau$ and generate a semantic grasp plan.
This produces a natural language description of the grasp position $\ell_p$ (e.g., ``grab the handle'') and approach orientation $\ell_R$ (e.g., ``grasp the handle from the side and keep clear of the pouring path'').

\p{Proposing Grasp Positions}
We next use the target mesh $\hat{g}_1$ to convert these language descriptions into localized grasp proposals.
We render $\hat{g}_1$ from $V$ different viewpoints, producing a multi-view set of 2D images
$\mathcal{I} = \{\mathbf{I}_1^{(i)}\}_{i=1}^{V}$.
We then input $\mathcal{I}$ and the semantic grasp position $\ell_p$ to a VLM, which outputs $V$ candidate grasp locations in pixel space:
\begin{equation} \label{eq:M2}
    \left\{\mathbf{u}_{\mathrm{VLM}}^{(i)} \right\}_{i=1}^{V} = \mathcal{F}_{\mathrm{VLM}}(\mathcal{I}, \ell_p)
\end{equation}
Here $\mathbf{u}_{\mathrm{VLM}}^{(i)} \in \mathbb{R}^{2}$ is the labeled pixel location for a grasp on the $i$-th image (e.g., the $x$-$y$ pixel coordinate).
These pixel coordinates are subsequently back-projected using their corresponding depth values and camera parameters to obtain grasp positions in the robot's 3D space:
\begin{equation} \label{eq:M3}
    \widetilde{\mathbf{P}}_{\mathrm{VLM}}^{(i)} = \Pi_i^{-1} \left(\mathbf{u}_{\mathrm{VLM}}^{(i)}, Z^{(i)}\right), \quad \widetilde{\mathbf{P}}_{\mathrm{VLM}}^{(i)} \in \mathbb{R}^{3}
\end{equation}
In \eq{M3}, $Z^{(i)}$ is the depth map associated with the $i$-th view, and $\Pi_i^{-1}$ denotes the back-projection operation defined by the corresponding camera parameters.
At this point we have a set of $V$ different grasp positions that the VLM has identified across the target object.
We condense this set by clustering points based on their density \cite{campello2015hierarchical}, grouping similar candidates while pruning isolated predictions.
Let $\{\mathcal{C}_i\}_{i=1}^{K}$ denote the $K$ resulting clusters.
We compute the centroid of each cluster to obtain
the grasp-point proposals:
\begin{equation} \label{eq:M4}
    \mathbf{P}_{\mathrm{VLM}}^{(i)} = \frac{1}{|\mathcal{C}_i|} \sum_{\widetilde{\mathbf{P}} \in \mathcal{C}_i} \widetilde{\mathbf{P}}, \quad \mathbf{P}_{\mathrm{VLM}}^{(i)} \in \mathbb{R}^{3}
\end{equation}
In summary: each point $\mathbf{P}_{\mathrm{VLM}}^{(i)}$ is a location where the VLM thinks the robot can perform a task-oriented grasp.
We hypothesize that this approach has advantages over other VLM grasp techniques since we are leveraging the digital twin to render \textit{multiple object images}, and then clustering and averaging the VLM proposals across that image set.
By contrast, methods like~\cite{deshpande2025graspmolmo} rely on a \textit{single view} of the object to generate a VLM-guided grasp proposal.

\p{Proposing Grasp Orientation}
In addition to identifying task-oriented grasp positions, we must also find the right rotation for aligning the gripper.
Because we will perform downstream optimization to fine-tune these poses, we do not need the gripper orientation to be perfect; instead, we just need to know \textit{which side of the object} to approach.
Similar to \cite{chen2024vlmimic}, we augment the rendered views of the target mesh $g_1$ with colored gripper wireframes aligned with the normals of its bounding box.
We denote the set of candidate wireframes by $\mathcal{W} = \{\mathbf{w}^{(i)}\}_{i=1}^{J}$, where each wireframe
$\mathbf{w}^{(i)}$ is associated with a candidate gripper orientation $\mathbf{R}^{(i)} \in \mathrm{SO}(3)$.
Similar to our approach for grasp positions, we visually prompt the VLM with $\mathcal{W}$ and the semantic orientation description $\ell_R$, and the VLM outputs a single orientation proposal:
\begin{equation} \label{eq:M5}
    \mathbf{R}_{\mathrm{VLM}} = \mathcal{F}_{\mathrm{VLM}} \left( \mathcal{W}, \ell_R, \ell_p \right), \quad \mathbf{R}_{\mathrm{VLM}} \in \mathrm{SO}(3)
\end{equation}
In \eq{M5} we also condition the VLM on $\ell_p$ to provide additional context about the part of the object that should be grasped.
Overall, $\mathbf{P}_{\mathrm{VLM}}^{(i)}$ and $\mathbf{R}_{\mathrm{VLM}}$ provide semantically meaningful grasp proposals, but these proposals are based on VLM reasoning and may not be grounded in the robot or environment.
We therefore treat these proposals as \textit{initial guesses} for the optimization in the following subsection.

\subsection{Optimizing the Proposed Grasps} \label{sec:M3}

The core of our approach is optimization: to convert the VLM-proposed positions and orientations into feasible and robust grasps, we refine these poses within our digital twin with domain randomization.
This optimization accounts not just for grasp success, but also for joint limits, collisions, object interactions, and other robot or environmental constraints.
We recognize that grasp success is sparse and non-convex.
We therefore leverage gradient-free Bayesian optimization, where the optimization is seeded by the task-aware proposals from Section~\ref{sec:M2}, and the optimization region is constrained to poses near these seeds.

\p{Optimization Region}
We are optimizing for the grasp pose $\mathbf{T} \in SE(3)$ from \eq{P2}.
This grasp pose is defined by a position-orientation pair, which we write as $\mathbf{T} = \left(\mathbf{P}, \mathbf{R}\right)$.
The initial candidate set of task-oriented grasp poses is:
\begin{equation} \label{eq:M6}
    \mathcal{T}_0 = \left\{\mathbf{T}_0^{(i)}\right\}_{i = 1}^K, \quad \mathbf{T}_0^{(i)} = \left(\mathbf{P}_{\mathrm{VLM}}^{(i)}, \mathbf{R}_{\mathrm{VLM}}\right)
\end{equation}
We constrain our optimization to regions around these poses.
This constraint ensures that the resulting grasp respects the high-level semantic meaning (i.e., performing the right grasp), while allowing for low-level adjustments to maximize effectiveness.
For each of the $i = 1, \ldots, K$ grasp proposals in \eq{M6}, we define the local search region as:
\begin{equation} \label{eq:M7}
    \mathcal{T}_{\mathrm{local}}^{(i)} =  \left\{ (\mathbf{P},\mathbf{R}) \;\middle|\; \begin{aligned}
        \left\| \mathbf{P}-\mathbf{P}_{\mathrm{VLM}}^{(i)} \right\|_{2} &\leq \epsilon_p \\
        d \left( \mathbf{R},\mathbf{R}_{\mathrm{VLM}} \right) &\leq \epsilon_R
    \end{aligned}
    \right\}
\end{equation}
where $\epsilon_p \geq 0$ and $\epsilon_R \geq 0$ specify the positional width and angular search radii, and $d$ denotes the angular distance between two rotation matrices.
The complete optimization
search space is the union of the local regions defined by \eq{M7}.

\p{Cost Function}
Now that we have the search space defined, we need to specify the cost function to optimize.
Since we are focusing on grasping, we want to make sure that the pose $\mathbf{T}$ results in a grasp that \textit{holds the target object}.
But because we are performing this optimization in a digital twin, we can also simulate whether reaching pose $\mathbf{T}$ is \textit{kinematically feasible} or results in \textit{unintended collisions} \cite{sundaralingam2023curobo}.

Let $J_{\mathrm{sim}}(\mathbf{T}, \Delta \theta)$ be the cost returned by the digital twin.
Here $\Delta \theta$ is a perturbation to the object poses ($xy$ position and yaw) in \eq{M1} for domain randomization.
During a simulation the robot plans a collision-free inverse kinematics to reach $\mathbf{T}$, closes its gripper, and then lifts up vertically.
Poses that are kinematically unreachable by the motion planner (e.g., paths with collisions, or paths that go beyond joint limits) are assigned a cost of 0.
The simulation cost $J_{\mathrm{sim}}(\mathbf{T}, \Delta \theta)$ returns $1$ if the object is successfully lifted at the end of the simulation; otherwise $J_{\mathrm{sim}}(\mathbf{T}, \Delta \theta) = 0$.

As mentioned above, the digital twin is only an \textit{estimate} of the environment; our knowledge of the objects' dynamic parameters is limited, and their positions and rotations may be slightly off. 
To make the optimization result more robust during sim-to-real transfer, we test a candidate pose $\mathbf{T}$ under a variety of perturbations $\Delta \theta$ to object $x$-$y$ position and rotation around $z$.
Let $p(\Delta\theta)$ be the distribution used for domain randomization.
Then the expected cost of a grasp is:
\begin{equation} \label{eq:M8}
    J(\mathbf{T}) = -\mathbb{E}_{\Delta \theta \sim p(\cdot)} \left[ J_{\mathrm{sim}}(\mathbf{T}, \Delta \theta) \right]
\end{equation}
Because $J_{\mathrm{sim}}$ is binary, $J(\mathbf{T})$ corresponds to the probability of success under the modeled perturbation distribution. 
We approximate this expectation using $M$ randomized simulation rollouts to reach our final cost function:
\begin{equation} \label{eq:M9}
    \widehat{J}(\mathbf{T}) = -\frac{1}{M} \sum_{i=1}^{M} J_{\mathrm{sim}} \left( \mathbf{T}, \Delta\theta^{(i)} \right), \quad \Delta\theta^{(i)} \sim p( \cdot )
\end{equation}
Note that a negative sign is included because we want to minimize this cost, which corresponds to maximizing the number of successful, collision-free grasps.

\p{Bayesian Optimization}
We perform batched Bayesian optimization to minimize \eq{M9} across a search space defined by the union of \eq{M7}.
Bayesian optimization was chosen because the underlying problem is non-convex with sparse rewards.
Let $\mathcal{D} = \{(\mathbf{T}_1, \widehat{J}_1), (\mathbf{T}_2, \widehat{J}_2), \ldots \}$ be the poses that we have previously tested and the costs associated with those poses: at the start of the optimization $\mathcal{D}$ is an empty set.
At the $t$-th round of optimization we leverage Thompson sampling to draw a batch of $B$ candidate poses: $\{\mathbf{T}_t^{(i)}\}_{i=1}^B$.
Thompson sampling leverages $\mathcal{D}$ to update the posterior used for drawing each batch; intuitively, this means it is more likely to sample poses that are close to ones that were successful in previous rounds of optimization.
The selected candidates are evaluated in parallel within the digital twin, and the resulting $(\mathbf{T}, \widehat{J})$ pairs are appended to $\mathcal{D}$.
After a total of $T$ rounds of optimization, we return the evaluated grasp pose that had the lowest cost within the digital twin:
\begin{equation} \label{eq:M10}
    \mathbf{T}^{*} = \left( \mathbf{P}^{*}, \mathbf{R}^{*} \right) = \underset{ (\mathbf{T},\widehat{J}\,)\in\mathcal{D} }{ \operatorname*{arg\,min} } \; \widehat{J}(\mathbf{T})
\end{equation}
No gradients are required for this approach.
We perform the Bayesian optimization in batches to accelerate the end-to-end process and test multiple proposals at once; batching is practical, but not theoretically necessary for convergence.

\p{Sim-to-Real}
The final step in our real-to-sim-to-real pipeline is taking the optimized grasp pose $\mathbf{T}^{*}$ and rolling it out on the real robot.
Since the robot in our digital twin matches our real-world system, we perform this transfer by directly using $\mathbf{T}^{*}$.
Given current state $s$, the robot plans a collision-free motion from $s$ to $\mathbf{T}^{*}$ \cite{sundaralingam2023curobo}, and then controls its joints to move along the waypoints of this plan.
The robot finally closes its gripper to grasp the target object.

\section{Experiments} \label{sec:experiments}

\begin{table}[t]
\vspace*{0.5em}
\footnotesize
\caption{Tasks used in our experiments. Most tasks were taken from related works \textbf{A}~\cite{rashid2023language}, \textbf{B}~\cite{tang2025affordgrasp}, and \textbf{C}~\cite{tang2025foundationgrasp}. All 30 tasks were used in our simulated testing (Table~\ref{table:table2}), and a subset of 10 tasks was used for real world evaluations (\fig{objective}).}
\label{table:table1}
\centering
\begin{tabular}{clccc}
\hline \# & Task & Source & Sim & Real \bigstrut \\ \hline
\rowcolor{gray!5} 1 & Hand me the knife & C & \cmark & \cmark \\
\rowcolor{gray!0} 2 & Cut the bread & A,B,C & \cmark & \cmark \\
\rowcolor{gray!5} 3 & Grab the mug by the handle & A,B,C & \cmark & \cmark \\
\rowcolor{gray!0} 4 & Flip the pancake with the spatula & B,C & \cmark & \cmark \\
\rowcolor{gray!5} 5 & Squeeze the mustard out the bottle & C & \cmark & \cmark \\
\rowcolor{gray!0} 6 & Grab a wet wipe from the pack & A & \cmark & \cmark \\
\rowcolor{gray!5} 7 & Close the box of wet wipes & A & \cmark & \cmark \\
\rowcolor{gray!0} 8 & Deliver the wine glass & A,B & \cmark & \cmark \\
\rowcolor{gray!5} 9 & Uncork the wine & A & \cmark & \cmark \\
\rowcolor{gray!0} 10 & Wear the sunglasses & A & \cmark & \cmark \\
\rowcolor{gray!5} 11 & Grab the spoon & A,B,C & \cmark &  \\
\rowcolor{gray!0} 12 & Grab the hot kettle & B & \cmark &  \\
\rowcolor{gray!0} 13 & Grab the saucepan & A,B,C & \cmark & \\
\rowcolor{gray!5} 14 & Open the saucepan & A & \cmark &  \\
\rowcolor{gray!0} 15 & Dispense the soup with a ladle & C & \cmark &  \\
\rowcolor{gray!5} 16 & Drink the Coke can & C & \cmark &  \\
\rowcolor{gray!0} 17 & Pour out the can & Ours & \cmark &  \\
\rowcolor{gray!5} 18 & Scrub the dishes & A,C & \cmark & \\
\rowcolor{gray!0} 19 & Grab the cleaning spray & A & \cmark & \\
\rowcolor{gray!5} 20 & Open the cleaning spray & A & \cmark & \\
\rowcolor{gray!0} 21 & Spray the table & A & \cmark & \\
\rowcolor{gray!5} 22 & Hand me the scissors & Ours & \cmark & \\
\rowcolor{gray!0} 23 & Cut the paper & B & \cmark & \\
\rowcolor{gray!5} 24 & Give me the screwdriver & A & \cmark & \\
\rowcolor{gray!0} 25 & Adjust the torque on the drill & Ours & \cmark & \\
\rowcolor{gray!5} 26 & Swing the hammer & A,B,C & \cmark & \\
\rowcolor{gray!0} 27 & Use the pliers & C & \cmark & \\
\rowcolor{gray!5} 28 & Grab the measuring tape & A & \cmark & \\
\rowcolor{gray!0} 29 & Pick up the book  & C & \cmark & \\
\rowcolor{gray!5} 30 & Give me the rose without the vase & A & \cmark & \\ \hline
\end{tabular}
\vspace{-1.5em}
\end{table}

\begin{table*}[t]
\vspace*{1em}
\caption{Grasp performance in simulated tasks from Table~\ref{table:table1}. Results are averaged over 30 tasks, each evaluated across 3 base scenes with five object configurations per scene. This yields 450 grasp attempts per method. The values are grasp success \% in the form $avg \pm sem$. All pairwise comparisons between GraspTwin and the baselines are statistically significant ($p< .05$).
}
\label{table:table2}
\centering
\begin{tabular}{cccccc}
\hline Metric [\%] & LERF-TOGO \cite{rashid2023language} & GraspMolmo \cite{deshpande2025graspmolmo} & Ours-O & Ours-G & GraspTwin \bigstrut \\ \hline
\rowcolor{gray!5} \textit{Task-Oriented} & $11.3 \pm 1.5$ & $53.1 \pm 2.4$ & $75.8 \pm 2.0$ & $51.3 \pm 2.4$ & $\textbf{81.1} \pm 1.8$ \\
\textit{Lift Object} & $9.1 \pm 1.4$ & $48.0 \pm 2.4$ & $28.9 \pm 2.1$ & $48.0 \pm 2.4$ & $\textbf{54.4} \pm 2.3$ \\
\rowcolor{gray!5} \textit{Collision Free} & $35.6 \pm 2.3$ & $76.4 \pm 2.0$ & $63.8 \pm 2.3$ & $87.3 \pm 1.6$ & $\textbf{95.6} \pm 1.0$ \\
\textbf{Overall} & $5.8 \pm 1.1$ & $33.1 \pm 2.2$ & $22.9 \pm 2.0$ & $33.8 \pm 2.2$ & $\textbf{51.6} \pm 2.4$ \\ \hline
\end{tabular}
\end{table*}

\begin{figure*}[t]
    \centering
    \includegraphics[width=1.0\linewidth]{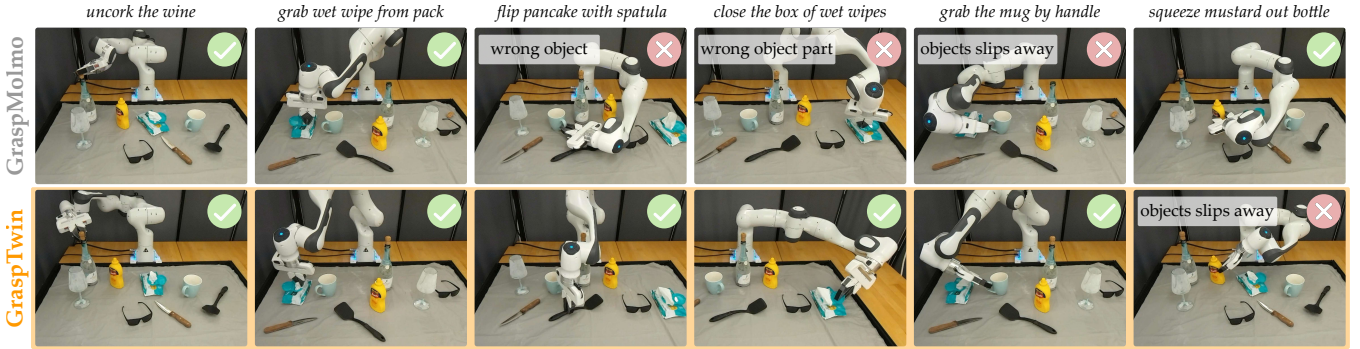}
    \caption{Examples of real-world grasps performed using GraspMolmo \cite{deshpande2025graspmolmo} (top row) and GraspTwin (bottom row). Each column shows a different task.}
    \label{fig:examples}
    \vspace{-1.5em}
\end{figure*}

\begin{figure}[t]
    \centering
    \vspace{0.5em}
    \includegraphics[width=1.0\linewidth]{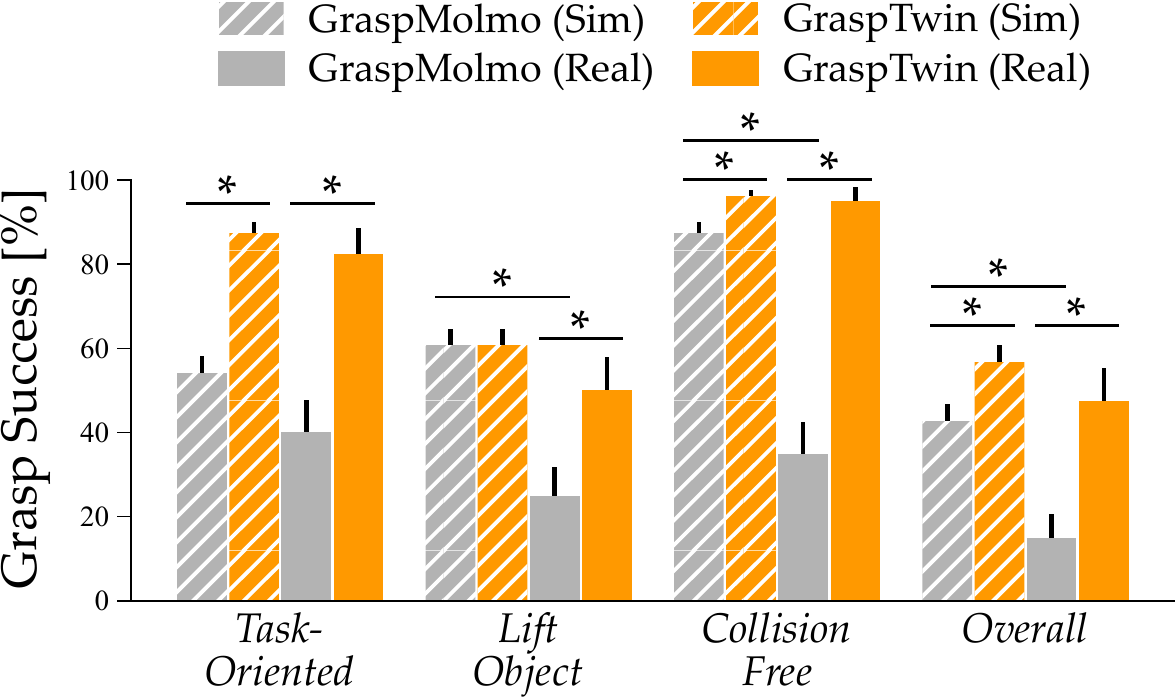}
    \caption{Grasp performance in the real-world tasks from Table~\ref{table:table1}. Results are averaged over 10 tasks with 4 trials per method. Error bars show standard error, and an $*$ denotes statistical significance ($p < .05$). The solid bars show real-world evaluation, and the hashed bars show performance on the same subset of tasks in simulation. We find that GraspTwin significantly outperforms GraspMolmo in real-world grasp success, and that the gap between GraspTwin's simulated and real-world performance is not significant.}
    \label{fig:objective}
    \vspace{-2em}
\end{figure}

We conduct simulated and real-world experiments to explore our proposed pipeline for task-oriented grasps.
Within these experiments we place a variety of objects on a table across 45 curated scenes, and give the robot a natural language task instruction (see Table~\ref{table:table1}).
The robot then tries to identify and perform a grasp that facilitates its task.
We separate our experiments into three parts.
In the first, we use \textit{simulations} to test 450 grasps across 5 different state-of-the-art grasp approaches.
In the second, we test our pipeline along with the best performing baseline in a set of \textit{real-world} experiments.
Finally, we report the performance of \textit{ablations} of GraspTwin that test its individual components. 

\p{Tasks and Environments}
Using SAM3D~\cite{sam3dteam2025sam3d3dfyimages}, 45 unique simulated scenes were created. 
Each scene contained 8 objects and a list of scene-specific tasks. 
To test the capability of these methods, 30 tasks of varying difficulty were chosen or created. 
The majority of the tasks were derived from well-known methods such as FoundationGrasp~\cite{tang2025foundationgrasp}, LERF-TOGO~\cite{rashid2023language}, and AffordGrasp~\cite{tang2025affordgrasp} as listed in Table~\ref{table:table1}.

\p{Independent Variables}
We tested a total of five different grasp pipelines. 
These included our method (\textbf{GraspTwin}), \textbf{GraspMolmo}~\cite{deshpande2025graspmolmo}, \textbf{LERF-TOGO}~\cite{rashid2023language}, and two ablations of our method.
The first ablation (\textbf{Ours-O}) evaluated the contribution of our optimization by deploying the raw VLM-proposed grasp pose without any fine-tuning in the digital twin. 
The second ablation (\textbf{Ours-G}) tested the importance of our task-oriented VLM proposal by instead using the pose from GraspMolmo as the seed to our optimizer. 

\p{Dependent Variables}
We defined a successful grasp based on three metrics: did the grasp lift and hold the object (\textit{Lift Object}), was the pose deployable without collisions with non-target objects (\textit{Collision-Free}), and was the grasp region semantically meaningful to the task (\textit{Task-Oriented}). 

\p{Experimental Setup}
We used a 7-DoF Franka Emika Panda robot arm in our simulations and real-world experiments. 
Both the simulation and the real-world setup were equipped with a UMI gripper~\cite{chi2024universalmanipulationinterfaceinthewild}. 
The algorithms in our experiments were run using two NVIDIA L40S GPUs.

\subsection{Comparing to Baselines in Simulation} \label{sec:E1}

We used simulations to test a wider range of environments, tasks, and methods.
For each of the 30 tasks in Table~\ref{table:table1} we set up 9 environmental layouts (e.g., different objects in the scene) and 5 runs (that randomized the object starting pose). Each task was deployed in 3 separate environmental layouts, allowing for 15 attempts per task. 

\p{Results}
Our results are summarized in Table~\ref{table:table2}.
We compare our method with LERF-TOGO (an approach relying on foundation models) and GraspMolmo (a trained end-to-end model).
The results show that GraspTwin outperforms the best baseline by 18.5\% (51.6\% vs. 33.1\% compared to GraspMolmo) in overall grasp success rate. 
Specifically, we see improvements in (a) selecting the task-oriented part of the object and (b) finding collision-free grasps.
The grasp robustness (i.e., lifting the object) remains comparable between our method and GraspMolmo.

Note that the overall grasp success is \textit{not an average} of these individual components: one grasp might fail to lift the object, while another fails to grasp the correct part.

\subsection{Testing Zero-Shot Transfer in Real-World Experiments} \label{sec:E2}

Next, we wanted to see whether these simulated results would transfer to a real-world robot platform.
We performed the 10 tasks marked in Table~\ref{table:table1} four times each with GraspTwin and the best performing baseline (GraspMolmo).
When performing real-world grasps we were particularly interested in zero-shot capabilities: would the proposed grasp work on its first attempt in the real setting?
Our experiments here follow the same procedure as outlined for Section~\ref{sec:E1}, but we limited the scope to only 40 trials.
Examples of successful and failed grasps are shown in \fig{examples}.

\p{Results}
Figure~\ref{fig:objective} summarizes the results.
We observe that our method transfers accurately to a real-world setting, only varying by 9.2\% in overall grasp success. 
This result supports our real-to-sim-to-real pipeline: the digital twin is sufficiently accurate to refine grasps for real-world execution.
Interestingly, we observe that GraspMolmo suffers a more significant sim-to-real performance gap, losing 27.7\% in its overall grasp success. 
We hypothesize that this is because GraspMolmo lacks the ability to fine-tune its proposals, and so it is heavily reliant on the quality of its calibration and sensing.
For example, real-world errors in depth estimation can lead to inaccurate grasp poses with GraspMolmo.

Comparing GraspTwin and GraspMolmo, our approach achieves roughly a \textit{three-fold performance improvement}.
The two components with the largest differences are semantic alignment, where GraspTwin finds task-oriented grasps twice as often as GraspMolmo, and obstacle avoidance, where GraspTwin finds collision-free grasp poses more than twice as often as GraspMolmo.

\vspace{-0.5em}
\subsection{Evaluating GraspTwin Components through Ablations} \label{sec:E3}

In the final part of our experiments we tested how each component of GraspTwin contributes to its success.
More specifically, we tested GraspTwin without the optimization process (Ours-O) and without our semantic grasp proposal approach (Ours-G).
We highlight that Ours-G is equivalent to using GraspMolmo to seed our optimization pipeline.

\p{Without Optimization} Table~\ref{table:table2} demonstrates that our method without the optimization process results in a significant reduction in overall task performance (51.6\% to 22.9\%). 
Removing the optimization (Ours-O) reduces the robustness (\textit{Lift Object}) and physical feasibility (\textit{Collision-Free}).
These results are in line with our formulation: the VLM generates a task-oriented grasp, but lacks the grounding for successful embodiment. 
As expected, the subsequent optimization process converts the semantic proposals into physically feasible grasps that can be executed on the robot. 

\p{Without Our Task-Oriented Proposals}
Removing our semantic grasp proposal and replacing it with GraspMolmo (Ours-G) sees a large decrease in \textit{Task-Oriented} grasps (e.g., picking the wrong object part).
This suggests that using the digital twin to generate multiple object views --- and then querying the VLM with these images --- offers a semantically meaningful initial guess for downstream optimization.
Interestingly, adding our optimization approach to GraspMolmo does not significantly increase its overall performance (comparing GraspMolmo to Ours-G).
The combination does increase the rate of collision-free grasps, showing that even end-to-end trained grasping foundation models (like GraspMolmo) may struggle grounding their proposals in the embodiment of the robot and the object itself, despite large datasets and specialized training.
\section{Conclusion} \label{sec:conclusion}

In this paper we introduce GraspTwin, a real-to-sim-to-real framework for zero-shot grasping.
Our framework tackles two key challenges: grasping the object in a way that facilitates the task, and making sure the grasp is actually grounded in the robot and environment.
By constructing a digital twin from an image of the robot's scene, we are able to render the target object from multiple perspectives and query VLMs for semantically meaningful grasp poses.
We then leverage these poses as the initial guesses for local optimization.
This optimization is performed within the digital twin: we roll out parallel tests with domain randomization to refine the proposed grasp poses and identify a physically feasible, collision-free, and robust solution.
GraspTwin outperformed state-of-the-art baselines in simulations (increasing success by 18 percentage points compared to the closest alternative) and real-world experiments (33\% absolute improvement).

\p{Limitations}
We see GraspTwin as a step towards robots that plan robust and task-oriented grasps on the fly.
One key limitation of our current framework is the time required to obtain a grasp.
In our experiments, the end-to-end GraspTwin pipeline took roughly 5 minutes.
The majority of this time was spent evaluating batched grasp proposals in the physics simulator.
Because these simulated rollouts can be executed concurrently, we anticipate that GraspTwin's execution time can be significantly reduced by scaling the computational hardware to increase batch parallelism.

\balance
\bibliographystyle{IEEEtran}
\bibliography{Bibtex}

\end{document}